\documentclass[lettersize,journal]{IEEEtran}
\usepackage{amsmath,amsfonts}
\usepackage{algorithmic}
\usepackage{algorithm}
\usepackage{array}
\usepackage[caption=false,font=normalsize,labelfont=sf,textfont=sf]{subfig}
\usepackage{textcomp}
\usepackage{stfloats}
\usepackage{url}
\usepackage{verbatim}
\usepackage{graphicx}
\usepackage{cite}
\usepackage{pdfpages}
\usepackage{amsmath}
\usepackage{float}
\usepackage{booktabs}
\usepackage{multirow}
\begin{document}

\title{Deep Learning-based Cross-modal Reconstruction of Vehicle Target from Sparse 3D SAR Image}

\author{Da~Li,
	Guoqiang~Zhao,
	Chen~Yao, 
	Kaiqiang~Zhu, 
	Houjun~Sun,
	Jiacheng~Bao
	and Maokun~Li,~\IEEEmembership{Fellow,~IEEE}
	\thanks{Jiacheng~Bao is the corresponding author.}
}

\markboth{Journal of \LaTeX\ Class Files,~Vol.~14, No.~8, August~2021}
{Shell \MakeLowercase{\textit{et al.}}: A Sample Article Using IEEEtran.cls for IEEE Journals}


\maketitle

\begin{abstract}

Three-dimensional synthetic aperture radar (3D SAR) is an advanced active microwave imaging technology widely utilized in remote sensing area. To achieve high-resolution 3D imaging, 3D SAR requires observations from multiple azimuthal aspects and altitude baselines surrounding the target. However, constrained flight trajectories often lead to sparse observations, which degrade imaging quality, particularly for anisotropic man-made small targets, such as vehicles and aircraft. In the past, compressive sensing (CS) was the mainstream approach for sparse 3D SAR image reconstruction. More recently, deep learning (DL) has emerged as a powerful alternative, markedly boosting reconstruction quality and efficiency. However, existing DL-based methods typically rely solely on high-quality 3D SAR images as supervisory signals to train deep neural networks (DNNs). This unimodal learning paradigm prevents the integration of complementary information from other data modalities, which limits reconstruction performance and reduces target discriminability due to the inherent constraints of electromagnetic scattering. In this paper, we introduce cross-modal learning and propose a Cross-Modal 3D-SAR Reconstruction Network (CMAR-Net) for enhancing sparse 3D SAR images of vehicle targets by fusing optical information. Leveraging cross-modal supervision from 2D optical images and error propagation guaranteed by differentiable rendering, CMAR-Net achieves efficient training and reconstructs sparse 3D SAR images, which are derived from highly sparse-aspect observations, into visually structured 3D vehicle images. Trained exclusively on simulated data, CMAR-Net exhibits robust generalization to real-world data, outperforming state-of-the-art CS and DL methods in structural accuracy within a large-scale parking lot experiment involving numerous civilian vehicles, thereby demonstrating its strong practical applicability. Our work highlights the potential of cross-modal learning for 3D SAR image reconstruction and emphasizes its value in both practical deployment and SAR target interpretation.

\end{abstract}


\section{Introduction}
\label{introduction}

\IEEEPARstart{T}{hree}-dimensional synthetic aperture radar (3D SAR) provides detailed spatial representations of a target’s electromagnetic scattering characteristics in three dimensions, operating reliably under all-weather, all-day conditions, which makes it invaluable for applications such as detailed target interpretation, urban remote sensing, and environmental monitoring \cite{zhuSuperresolvingSARTomography2014,9520113,10856784}. Its core imaging mechanism involves acquiring radar data from multiple azimuths and altitude baselines, forming a synthetic aperture to achieve 3D resolution. However, this imaging mode inevitably faces a trade-off between image quality and acquisition cost. For anisotropic complex targets and scenes, obtaining clear and detailed 3D reconstructions generally requires dense observations across multiple azimuthal aspects and multiple altitude baselines to enhance imaging. However, in practical applications, multi-aspect multi-baseline observation is frequently hindered by data sparsity due to limitations in flight trajectories, terrain, and weather conditions. This results in degraded 3D reconstruction quality, particularly for anisotropic man-made small targets like vehicles and aircraft \cite{wangMultibaselineSAR3D2023,8010434}. To address this, sparse reconstruction algorithms are widely employed to mitigate quality degradation under sparse observations.

In typical sparse 3D SAR application scenarios, scatterers are often sparsely distributed along the elevation dimension, with only a few dominant scatterers located within each range-azimuth resolution cell\cite{10105605}. As a result, CS-based methods have long been the dominant approach for addressing the 3D SAR inversion problem\cite{austinSparseMultipass3D2009,zhuTomographicSARInversion2010}. These methods formulate the sparse imaging problem as a signal recovery model. By incorporating various penalty functions and optimization techniques, they can reconstruct high-resolution target images from incomplete measurements. However, CS-based approaches also suffer from several limitations. The iterative optimization process is computationally intensive\cite{yangCompressedSensingRadar2019a,potterSparsityCompressedSensing2010a}, and in 3D scenarios, the computational complexity grows exponentially with data dimensionality, rendering large-scale processing increasingly intractable. Moreover, CS solvers are inherently sensitive to noise, require careful hyperparameter tuning, and often lack robustness and generalization capability across diverse imaging conditions\cite{xuSparseSyntheticAperture2022}.

With the rapid advancement of DL technologies and their widespread adoption across various domains, DL-based sparse reconstruction methods have emerged as a promising solution to address the aforementioned limitations. DL is a data-driven approach that learns the underlying features of a scene directly from training data. By employing hierarchical model architectures, it captures complex nonlinear mappings between inputs and outputs. In recent years, numerous studies have explored the use of deep neural networks (DNNs) for tackling sparse imaging tasks\cite{wang3DSARDataDriven2022,sunLargescaleBuildingHeight2021,wangTomoSAR3DReconstruction2021a,sun3DRIMR3DReconstruction2021a}. These methods can generally be categorized into two main types.
\begin{figure*}[t]
	\centering
	\includegraphics[width=1\linewidth]{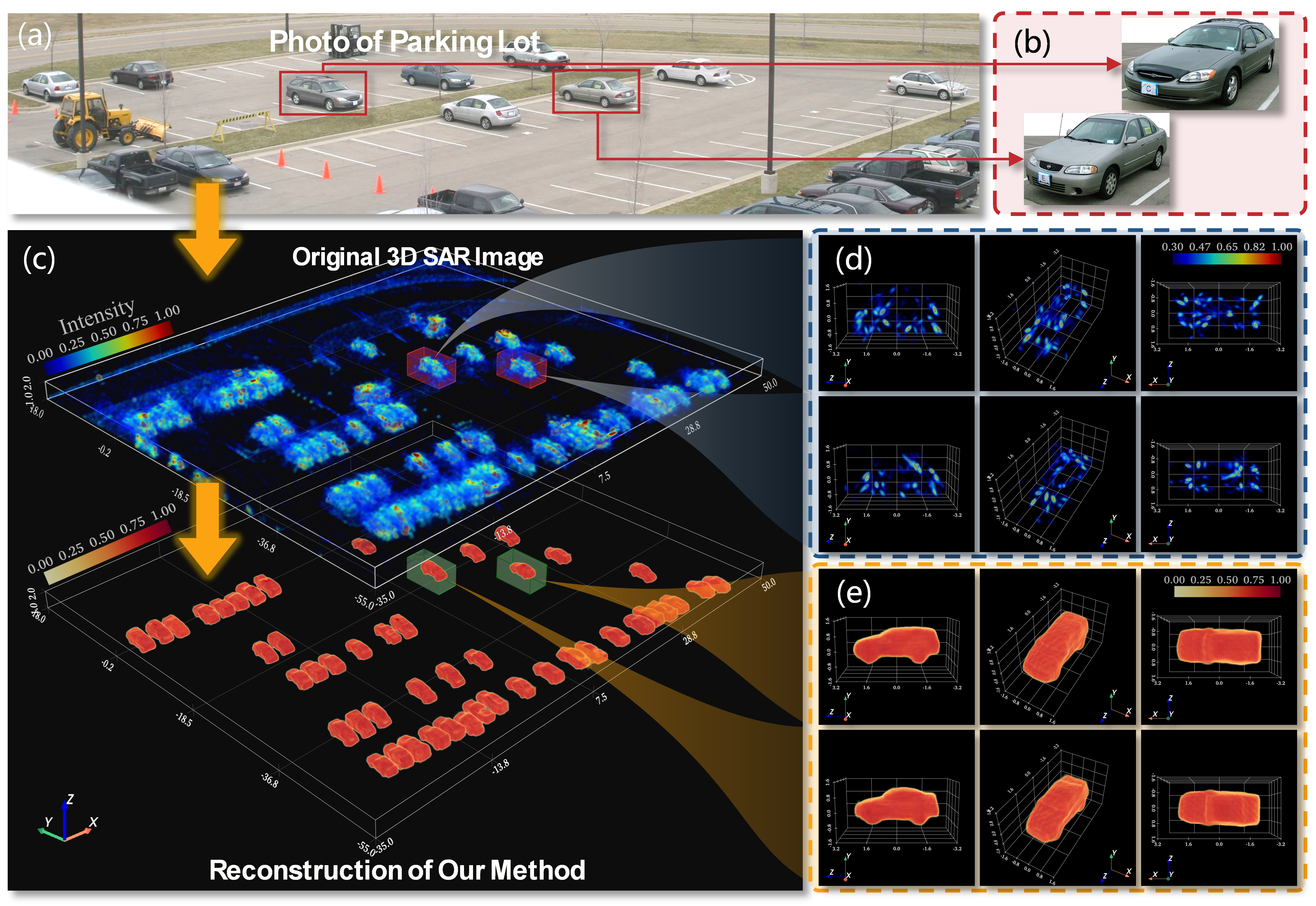}
	\caption{Reconstruction results of CMAR-Net on 48 real-measured vehicles in a parking lot. (a) A wide-view optical photo of the parking lot\cite{casteel2007challenge}. (b) Sample optical photos of individual vehicles. (c) Raw sparse-aspect multi-baseline SAR 3D image and corresponding reconstruction result of CMAR-Net. (d) and (e) Enlarged views of two selected vehicles for detailed comparison.}
	\label{fig:fullscene}
\end{figure*}

The first type embeds DNNs into the CS framework by replacing the traditional iterative optimization process with cascaded DNN modules that approximate its nonlinear components. These approaches are commonly referred to as unfolding methods\cite{10781458,wang3DSARAutofocusing2022,wangTPSSINetFastEnhanced2021a,zhouSAF3DNetUnsupervisedAMPInspired2022a}. Through extensive training, these networks learn data priors and optimization parameters, eliminating the need for manual configuration. Owing to their parallel architecture and inference-based (i.e., non-iterative) processing paradigm, these approaches require only a single forward pass to achieve reconstruction accuracy comparable to that of traditional CS algorithms, thereby significantly reducing computational complexity.

The second type approach employs a sequential architecture, cascading DNNs with traditional imaging algorithms in a "two-step" process. First, a conventional algorithm produces an initial 3D SAR image. This image is then refined by a DNN to enhance the imaging quality. For example, in \cite{wangSingleTargetSAR2021}, a 3D U-Net was utilized to improve back-projection (BP) results, enabling high-resolution reconstruction from distorted BP images. Similarly, in \cite{wangMultibaselineSAR3D2023}, a GAN-based sparse-aspects-completion network (SACNet) enhanced the results of CS imaging, demonstrating that a network trained on simulated data can still achieve effective target reconstruction on real-world data. These methods treat the improvement of 3D SAR imaging results as an image enhancement or denoising task. Leveraging the powerful data representation capabilities of DNNs and extensive training, they can rapidly and reliably reconstruct high-resolution 3D target images, even under conditions of highly sparse observation data.

In summary, DL-based sparse 3D SAR reconstruction methods demonstrate significant potential in enhancing both imaging efficiency and quality\cite{xuSparseSyntheticAperture2022}. However, current learning-based approaches still face significant practical problems.

\begin{enumerate}
	\item DL-based methods learn the mapping relationships between training data pairs, which are derived from the single modality (electromagnetic information). This reliance on electromagnetic constraints imposes inherent limits on the achievable resolution, restricts further improvements in image quality, and hinders image discriminability.
	\item As a data-driven technology, the performance of DL-based methods heavily depends on the quality of training data. Most existing frameworks utilize high-resolution 3D SAR images as supervision, which necessitates imaging preprocessing of full-observation data prior to training. This process demands substantial computational time and storage space, or high-performance real-time processing platforms.
\end{enumerate}

To address the aforementioned limitations—particularly the resolution constraints imposed by single-modality data—we introduce cross-modal learning in this work. The cross-modal learning explores the intrinsic relationships between different data modalities, seeks shared representations, and integrates heterogeneous information. This concept has recently been successfully applied to problems such as cross-modal retrieval\cite{10843094}, image captioning\cite{9706348}, and SAR-to-optical image translation in remote sensing\cite{LI202114,PAN2024258,9627647}. 

In this work, we propose a sparse 3D SAR reconstruction framework driven by cross-modal learning, implemented through a novel network, CMAR-Net. The network reconstructs raw sparse 3D SAR image of vehicle targets into visually structured and geometrically accurate 3D representation. Figure \ref{fig:fullscene} shows the reconstruction results of CMAR-Net on 48 real-measured vehicles in a parking lot. Unlike conventional methods that require 3D supervision, our network is trained using aligned 2D optical images as supervisory signals. The key challenge lies in bridging the dimensional gap between 2D pixels and 3D voxels. To resolve this, we adopt differentiable rendering to compute consistency errors that are back-propagatable through the network.

The main contributions of this work are summarized as follows:

\begin{enumerate}
	\item \textbf{First cross-modal learning framework for 3D SAR reconstruction:} To our knowledge, this is the first application of cross-modal learning to 3D SAR reconstruction. Our method achieves high-resolution reconstructions with complete visual structures and sharp features, overcoming the inherent resolution limitations of electromagnetic imaging.
	\item \textbf{Robust CMAR-Net architecture:} We design a novel network incorporating a data augmentation strategy and a projection-reprojection module to enhance robustness and generalization. Remarkably, CMAR-Net trained on simulated data achieves strong performance on real-world datasets without any fine-tuning.
	\item \textbf{Reduced supervision requirements:} The proposed method only necessitates 2D optical images as supervision, eliminating dependency on high-resolution full-aspect (360°) data. This significantly simplifies the creation of high-quality imaging datasets, broadening the practicality of 3D SAR reconstruction.
	\item \textbf{State-of-the-art performance:} Extensive experiments show our method excels under low SNR and highly sparse aspect, outperforming existing approaches by 75.83\% in PSNR and 47.85\% in SSIM.
	
\end{enumerate}

This paper is organized as follows: Section \ref{method} introduces the imaging geometry, overall framework of the proposed method, and module details. Section \ref{experiment} presents the experimental setup, results, and analysis. Section \ref{conclusion} concludes the work.

\section{Methodology}
\label{method}
\begin{figure}[t]
	\centering
	\includegraphics[width=0.95\linewidth]{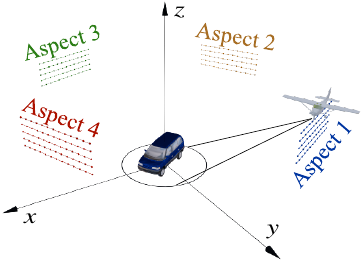}
	\caption{Multi-aspect multi-baseline SAR observation geometry.}
	\label{fig:geometry}
\end{figure}

\subsection{Multi-aspect Multi-baseline 3D SAR Observation Geometry}
The geometry of multi-aspect multi-baseline 3D SAR observation is illustrated in Figure \ref{fig:geometry}. In each observation aspect, the radar platform follows a straight flight path at different altitudes, forming observation apertures in both the azimuth and height dimensions, represented by evenly spaced points in the figure. To capture anisotropic scattering information and achieve precise 3D reconstruction, the radar platform must conduct multi-aspect observations around the target, measuring the scattering characteristics from various directions, with different colors used in the figure to differentiate the observation aspects. However, in practical applications, the available observation aspects are often sparse due to limitations in flight conditions, acquisition costs, or platform configurations. Such aspect sparsity not only reduces the diversity of scattering measurements but also degrades the angular resolution of anisotropic targets, ultimately impairing the overall imaging quality. Therefore, in this work, we specifically focus on aspect sparsity as the key challenge.

\subsection{Method framework}
\begin{figure*}[h]
	\centering
	\includegraphics[width=0.95\linewidth]{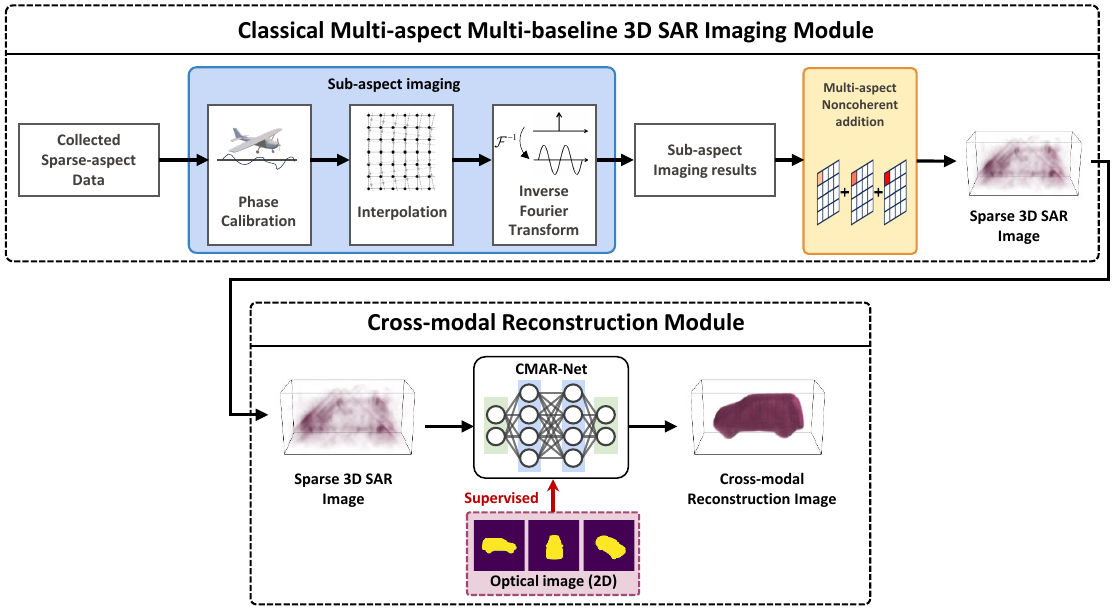}
	\caption{The framework of the proposed sparse 3D SAR cross-modal reconstruction method.}
	\label{fig:methodframework}
\end{figure*}
The framework of the proposed sparse 3D SAR reconstruction method is illustrated in Figure \ref{fig:methodframework}. It consists of two main processing modules: the classical multi-aspect multi-baseline 3D SAR imaging module and the cross-modal reconstruction module. First, classical 3D SAR imaging process is performed on the collected sparse-aspect, multi-baseline radar data. This step includes sub-aspect imaging and the non-coherent integration of multi-aspect images and can be regarded as a pre-imaging step. Subsequently, the proposed cross-modal reconstruction network, CMAR-Net, further enhances the 3D SAR image. As shown in the figure, using a vehicle target as an example, CMAR-Net directly takes the 3D SAR image as input and produces a clearer and more structurally complete 3D reconstruction. During training, optical images of vehicles—spatially aligned with the SAR images—are introduced as supervision signals. This allows for quantitative evaluation of the 3D reconstruction accuracy and establishes a cross-modal supervision mechanism, thereby enabling modality transfer from SAR to optical. The details of each module are explained below.

\subsection{Classical multi-aspect multi-baseline 3D SAR imaging module}
The detailed process of multi-aspect multi-baseline 3D SAR imaging is illustrated in the upper part of Figure \ref{fig:methodframework}. The target is observed from several aspects, and we begin by independently imaging each sub-aspect. For each sub-aspect, after motion compensation and phase calibration, the polar-format frequency domain data are interpolated onto a Cartesian grid, and a three-dimensional inverse Fourier transform is then applied to reconstruct the target's spatial-domain image\cite{austinSparseMultipass3D2009}. Finally, all sub-aspect images are non-coherent added to integrate multi-aspect scattering information and generate the final 3D SAR image of the target\cite{dungan2010civilian}.

\subsection{Cross-modal reconstruction module}
\begin{figure*}[h]
	\centering
	\includegraphics[width=1\linewidth]{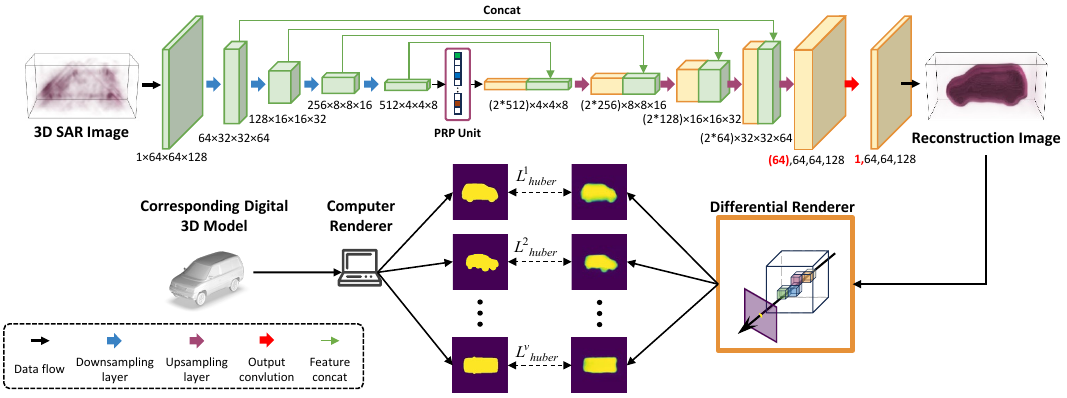}
	\caption{Network Architecture.}
	\label{fig:nework}
\end{figure*}

The cross-modal reconstruction module consists of a cross-modal reconstruction network, CMAR-Net, as shown in the lower part of Figure \ref{fig:methodframework}. It takes 3D SAR image data as input and utilizes spatially aligned multi-view 2D optical images as supervision signals. During training, the network minimizes the differences between the projected views of the 3D reconstruction result and the corresponding reference optical images from multiple viewpoints, encouraging the reconstructed 3D geometry to align accurately with the structures observed in the optical domain. The detailed architecture and training process of CMAR-Net are depicted in Figure \ref{fig:nework}. The following sections describe each component in detail.

\subsubsection{Network architecture}
The network consists of a contraction path (on the left), an expansion path (on the right), and a Projection-Reprojection (PRP) unit. The contraction path includes four downsampling layers, each consisting of a 3D convolutional layer followed by a max-pooling layer. The convolutional layers apply LeakyReLU activation and progressively increase the number of feature channels, while the pooling layers reduce the data dimensionality. The contraction path ends with a connection to the PRP unit.

The expansion path mirrors the contraction path, consisting of four upsampling layers and a single convolutional layer. Each upsampling layer uses a 3D transposed convolution with ReLU activation. Feature maps from the expansion path are combined with those from the contraction path at the same resolution through skip connections before being passed through the upsampling layers. These layers reduce the number of feature channels while increasing spatial resolution. The final upsampling layer produces a 64-channel output, then refined by a output convolution layer to smooth and integrate information across all channels, ensuring a more refined reconstruction.

\subsubsection{Projection-Reprojection unit}
The PRP unit inserted between the encoder and decoder is specially designed for our task. The differences between data modalities cause the encoder and decoder to process semantically distinct information at many low-level layers\cite{10214202}. To ensure that the network learns a common representation space between the two modalities, it is essential to design an encoding unit that is both expressive and resistant to overfitting.

\begin{figure}[t]
	\centering
	\includegraphics[width=1\linewidth]{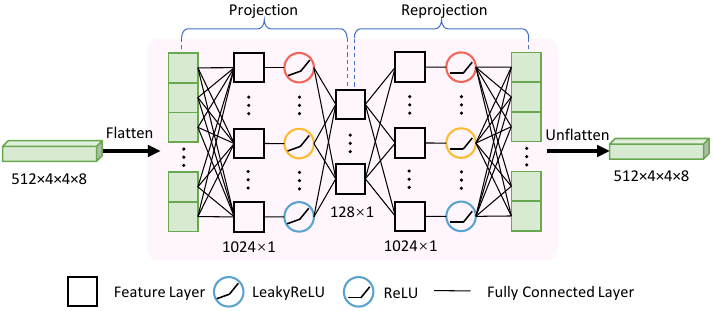}
	\caption{Architecture of PRP unit.}
	\label{fig:prp-unit}
\end{figure}

Therefore, inspired by \cite{10214202}, we design the PRP unit. The structure is detailed in Figure \ref{fig:prp-unit}. The projection part consists of two fully connected layers. The first fully connected layer, activated by LeakyReLU, compresses the input feature map into a low-dimensional vector. The second fully connected layer eliminates the activation and further compresses the feature's dimensionality to produce the latent representation $z$. The reprojection part reverses the projection process by symmetrically expanding the feature vector and reshaping it back to the original feature map size. The only difference is that the final fully connected layer uses ReLU activation to ensure consistency in the data dynamics within the network.

\subsection{Differentiable volume rendering}

To supervise the 3D reconstruction process using 2D optical images, a projection transformation technique is required to map volumetric 3D data onto a 2D image plane, thus enabling quantitative error evaluation between the two domains. However, to integrate this projection module into neural network training, the transformation process must remain differentiable to ensure gradient backpropagation of training errors. In CMAR-Net, we use differentiable volume rendering\cite{Niemeyer_2020_CVPR} to achieve this critical mapping.

Rendering refers to the computational process that converts mathematical representations of 3D scenes (e.g., geometric models, material properties, lighting conditions, and camera parameters) into 2D images. Differentiable volume rendering is a specialized variant for rendering 3D volumetric data(e.g. density fields). Its core principle involves integrating density and color values along light rays to establish a differentiable mapping from the 3D scene to the 2D image.

\begin{figure}[t]
	\centering
	\includegraphics[width=0.75\linewidth]{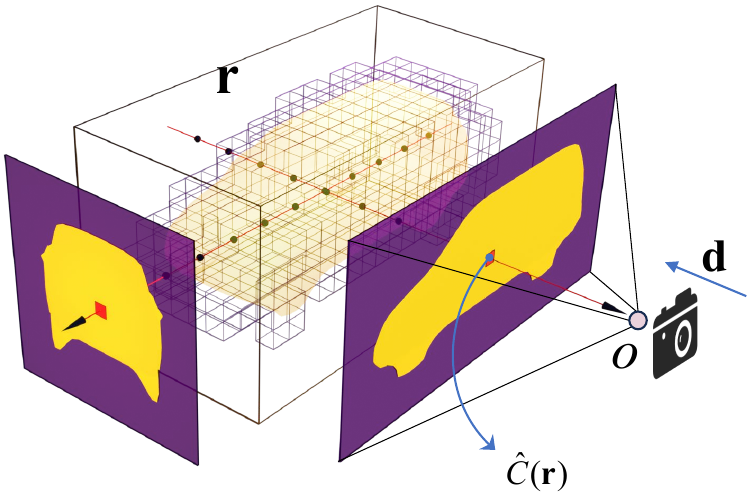}
	\caption{Rendering diagram.}
	\label{fig:rendering-diagram}
\end{figure}

Figure \ref{fig:rendering-diagram} illustrates the differentiable volume rendering process. Given a 3D volumetric data $\mathbf{V}\in \mathbb{R}^{W\times H\times D}$, a camera position $o$, and a viewing direction $\mathbf{d}$, volume rendering computes the pixel values $C(\mathbf{r})$ of a camera ray $\mathbf{r}(t) = o+t\mathbf{d}$ with near and far bounds $t_n$ and $t_f$ using the following formula:

\begin{equation}
	C({\bf{r}}) = \int_{{t_n}}^{{t_f}} T (t) \cdot \sigma ({\bf{r}}(t)) \cdot {\bf{c}}({\bf{r}}(t),{\bf{d}})dt.
\end{equation}
Here, $ \sigma (\mathbf{r}(t)) $ and $ {\bf{c}}({\bf{r}}(t),{\bf{d}}) $ represent the volume density and color at point $\mathbf{r}(t)$ along the camera ray with viewing direction $\mathbf{d}$, and $dt$ represents the step size of the ray at each integration step. $ T(t) $ is the cumulative transmittance, which indicates the probability that the ray propagates from $ t_n $ to $ t_f $ without being intercepted. The $T(t)$ is given by the following equation:

\begin{equation}
	T(t)=\text{exp}\left( -\int_{t_n}^{t} \sigma (\mathbf{r}(u))\cdot du \right).
\end{equation}

To compute the continuous integral, we discretize the ray by sampling it at equidistant depths and apply the integration rules discussed in the literature \cite{468400} to estimate the pixel value $ C(\mathbf{r}) $. This is given by the following formula:
\begin{equation}
	\hat{C}(\mathbf{r}) = \sum\limits_{i=1}^{N} T_i (1-\text{exp}(-\sigma_i \delta_i)){{\bf{c}}_i},
\end{equation}
where
\begin{equation}
	T_i = \text{exp} \left( -\sum\limits_{j=1}^{i-1} \sigma_j \delta_i \right).
\end{equation}

The $ \delta_i = t_{i+1} - t_{i}$ represents the distance between adjacent samples. The final rendered image is obtained by concatenating the estimated pixel values along all camera rays passing through the pixels of the rendering canvas, as expressed by:

\begin{equation}
	I=
	\left[
	\begin{array}{ccc}
		\hat{C}(\mathbf{r}_{0, 0}) & \dots & \hat{C}(\mathbf{r}_{w-1, 0}) \\
		\vdots & \ddots & \vdots \\
		\hat{C}(\mathbf{r}_{0, h-1}) & \dots & \hat{C}(\mathbf{r}_{w-1, h-1}) 
	\end{array}
	\right],        
\end{equation}
$ \mathbf{r}_{i,j} $ represents the camera ray passing through the pixel at $ (i, j) $, where $ i\in[0,h-1] $ and $ j\in[0,w-1] $, with $ h $ and $ w $ denoting the height and width of the rendered image, respectively.

In volume rendering models, each voxel location within 3D space typically encodes two fundamental attributes: volume density $\sigma$ and color $\bf{c}$. The density parameter quantifies a spatial position's light attenuation characteristics, where higher density values indicate greater probability of light occlusion. In several existing 3D reconstruction methods \cite{nerf, Chen_2023_ICCV, nerf3drec}, it is commonly assumed that regions with higher density are more likely to correspond to actual physical structures. As a result, density is often used as a proxy for the probability of object presence in space, and is subsequently employed to generate the 3D structural image of the object.

The electromagnetic nature of SAR makes it challenging to model color information. In our work, we focus on reconstructing the object’s geometric structure by using the volumetric density parameter as a proxy for the spatial occupancy probability. We simplify the rendering process by treating the color term $\bf{c}$ as a constant, effectively ignoring the influence of color. Meanwhile, to ensure consistent supervision, the optical images of vehicles used for training are preprocessed into binary masks that retain only geometric shape information. In these masks, pixels with a value of 1 indicate the presence of vehicle structures and serve as strong supervision signals for object existence in 3D space. This design encourages the network to assign higher density values to corresponding regions. Through sufficient training, the network learns the structural distribution of the target from these supervisory images and ultimately produces 3D reconstructions of vehicles that are spatially well-aligned with the optical supervision.

\subsection{Loss design}
We use mask images rendered from 3D vehicle models viewed from different angles as ground truth to compute the reconstruction error. During the training process, we employ the Huber loss function to quantify the error. The Huber loss between the $i$-th viewpoint's reconstructed image, $I_{\rm{r}}^i$, and the corresponding ground truth image, $I_{\rm{g}}^i$, is defined as:

\begin{equation}
	\mathcal{L}_{{\rm{huber}}}^i({I_{\rm{g}}^i},{I_{\rm{r}}^i},\gamma ) = 
	\begin{cases} 
		\frac{1}{2} \left( {I_{\rm{g}}^i} - {I_{\rm{r}}^i} \right)^2 & \text{if } \left| {I_{\rm{g}}^i} - {I_{\rm{r}}^i} \right| \leq \gamma, \\
		\gamma \left| {I_{\rm{g}}^i} - {I_{\rm{r}}^i} \right| - \frac{1}{2} \gamma^2 & \text{if } \left| {I_{\rm{g}}^i} - {I_{\rm{r}}^i} \right| > \gamma.
	\end{cases}
\end{equation}

Here, $\gamma$ is a adjustable parameter of the Huber loss. For $V$ rendering viewpoints, the total loss function is defined as their average:
\begin{equation}
	\mathcal{L} = \frac{1}{V}\sum\limits_{i = 0}^{V - 1} {\mathcal{L}_{{\rm{huber}}}^i({I_{\rm{g}}^i},{I_{\rm{r}}^i},\gamma )}. 
\end{equation}

\section{Experiments}
\label{experiment}
\subsection{Experimental Settings}
To train CMAR-Net, we construct a civilian vehicle dataset, including simulated sparse-aspect 3D SAR images and multi-view optical images. For validation, we perform extensive comparative experiments with baseline methods on the simulated data, including feature space interpolation. We then validate the generalization performance of CMAR-Net using real-world data and conduct ablation studies on the PRP unit and data augmentation strategy. This section provides a detailed overview of the experiment setup.

\subsubsection{Dataset and Augmentation}
\begin{figure*}[h]
	\centering
	\includegraphics[width=0.8\linewidth]{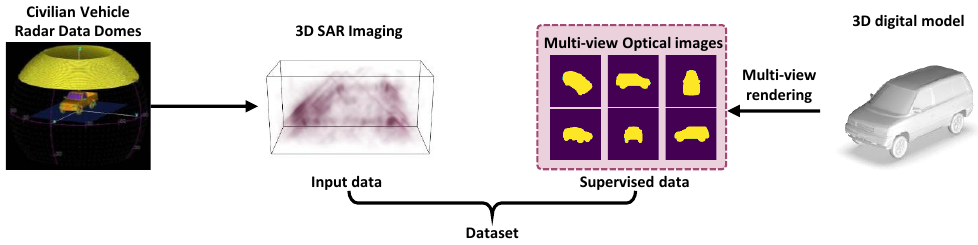}
	\caption{Dataset construction.}
	\label{fig:synthetic-dataset}
\end{figure*}

\begin{figure*}
	\centering
	\includegraphics[width=0.7\linewidth]{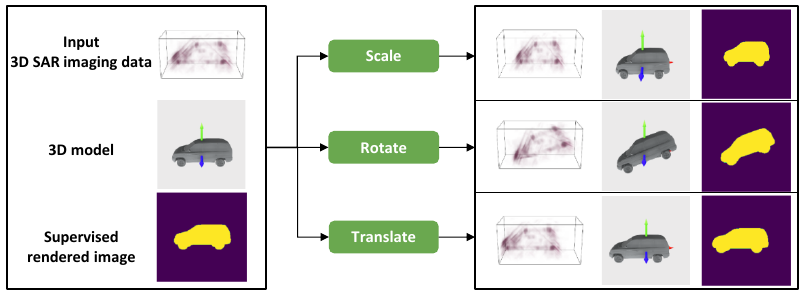}
	\caption{Data augmentation strategy.}
	\label{fig:augmentation}
\end{figure*}
The sparse-aspect multi-baseline 3D SAR images in the dataset were generated using the Civilian Vehicle Radar Data Domes (CVDomes), published by the United States Air Force Laboratory\cite{dungan2010civilian}. This data includes full-polarization, far-field X-band electromagnetic scattering data for ten civilian vehicles, covering a 360° azimuth and a pitch angle range from 30° to 60°. The viewpoint distribution and parameter settings of the simulation scene are illustrated in Figure \ref{fig:synthetic-dataset} and Table \ref{CV_Parameters}. For the training set, data from five vehicles—two sedans, two SUVs, and one pickup truck—are selected. Imaging is performed at eight pitch angles between 44.25° and 46°, with a spacing of 0.1875° between each angle. The 360° azimuth data are divided into 72 aspects, each 5° wide and spaced 0.0625° apart. Sub-aspect imaging is then performed, and finally, 8 randomly selected aspects are non-coherently summed to generate the 3D SAR images for training.
\begin{table}[t]
	\centering
	\caption{Simulated Parameters of the Civilian Vehicle Radar Data Domes and Training data setting\cite{dungan2010civilian}.}
	\label{CV_Parameters}
		\begin{tabular}{@{}cc@{}}
			\toprule[1.5pt]
			Parameter              & Value      \\ \midrule
			Radar center frequency & 9.6GHz     \\
			Unambiguous range      & $\approx 15m$      \\
			Extrapolation extent   & $\leq 0.25^\circ$     \\
			Azimuth extent         & $360^\circ$       \\
			Elevation extent       & $30^\circ - 60^\circ$ \\
			Angle extent for each aspect       & $5^\circ $ \\
			Total aspect number       &$ 72$ \\
			Elevation range for training       & $44.25^\circ - 46^\circ$ \\ \bottomrule[1.5pt]
		\end{tabular}%
\end{table}

The multi-view optical images used for supervision in the dataset are generated by rendering 3D digital models of the vehicles. We collect 3D models of the same vehicle types as those in the CVDomes and rendered optical binary images from multiple viewpoints. The process of assembling the dataset is illustrated in Figure \ref{fig:synthetic-dataset}.

Additionally, we design a data augmentation strategy to expand our dataset. As shown in Figure \ref{fig:augmentation}, the augmentation involves three operations: scaling $S$, rotation $R$, and translation $T$. For each pair of input raw 3D SAR image, a geometric transformation is applied, which is also performed on the corresponding 3D digital model. This transformation is then reflected in the rendered images. This strategy increases the sampling density of the data space, which enhances the network's generalization.

\subsubsection{Implementation}

In the classical multi-aspect multi-baseline 3D SAR imaging step, the imaging scene is set with dimensions of  $3.2m \times 3.2m \times 6.4m$ and a spatial resolution of $0.05m$, resulting in a data size of $64 \times 64 \times 128$. The 3D SAR data is then normalized to a dynamic range of [0, 1] before being input into the network. For the differentiable volume rendering module, the rendered image size is set to 256 $\times$ 256, with the camera positioned 7m from the scene center. The camera is fixed to point toward the center from 8 predetermined angles.

During training, the hyperparameters are set as follows: Huber loss parameter $\gamma=0.7$, batch size = 1, and the Adam optimizer with momentum parameters $\beta_1=0.5$ and $\beta_2=0.9$. The initial learning rate is set to $1\times10^{-4}$, gradually decreasing to $5\times10^{-5}$. The implementation is based on the PyTorch framework, with training and inference conducted on a machine equipped with an NVIDIA RTX A6000. The network is trained and evaluated on the simulated data, and, without any further fine-tuning, it is ultimately validated on the real-world data.

\subsubsection{Baseline methods}

To demonstrate the improvement in imaging quality achieved by integrating optical information, we use traditional imaging methods and DL methods using radar image supervision as baseline comparisons. The traditional methods include BP\cite{6721363} and CS\cite{ertinGOTCHAExperienceReport2007}, while the DL methods include SACNet\cite{wangMultibaselineSAR3D2023} and UNet3D\cite{wangSingleTargetSAR2021}, which are specifically designed to enhance the quality of sparse 3D SAR imagingand are trained using full-aspect 3D SAR images as supervision signals. In addition, we extend the backbone architectures of SACNet and UNet3D with cross-modal (\textbf{CM}) supervision, and use them as two additional baseline methods.

\subsubsection{Evaluation Metrics}
We convert the vehicle's 3D digital model into volume data with the same spatial extent and resolution as the imaging settings, using it as the ground truth. The imaging quality of our method is then evaluated using full-reference metrics, including Peak Signal-to-Noise Ratio (PSNR)\cite{5596999}, Structural Similarity Index Measure (SSIM)\cite{5596999}, Intersection over Union (IoU)\cite{wangMultibaselineSAR3D2023}, and Cross-Entropy (CE)\cite{wangMultibaselineSAR3D2023}.

\subsection{Results for Simulated Data}

\subsubsection{3D Reconstruction Results of CMAR-Net}
\begin{figure*}[!h]
	\centering
	\includegraphics[width=0.7\linewidth]{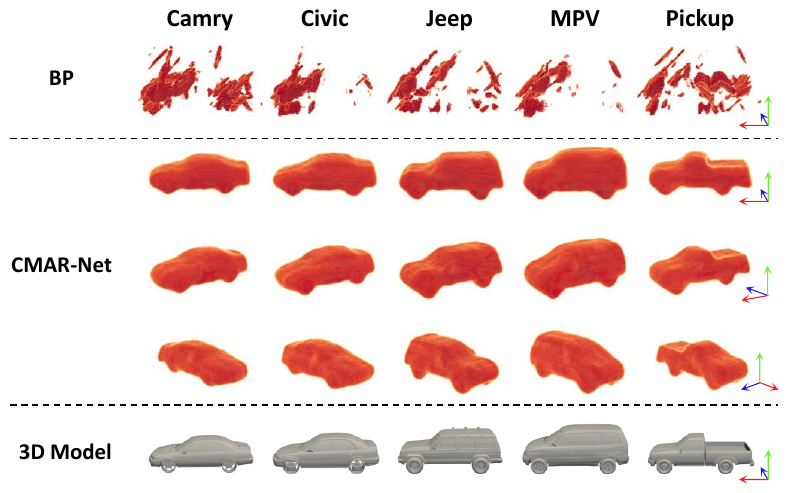}
	\caption{Overview of 3D reconstruction performance of CMAR-Net. BP results (1st row), CMAR-Net reconstructions (2nd-4th rows) and ground truth (5th row). Imaging at an SNR of \textbf{30dB} and \textbf{8} observation aspects.}
	\label{fig:simu1}
\end{figure*}
\begin{figure*}[!h]
	\centering
	\includegraphics[width=0.7\linewidth,page=1]{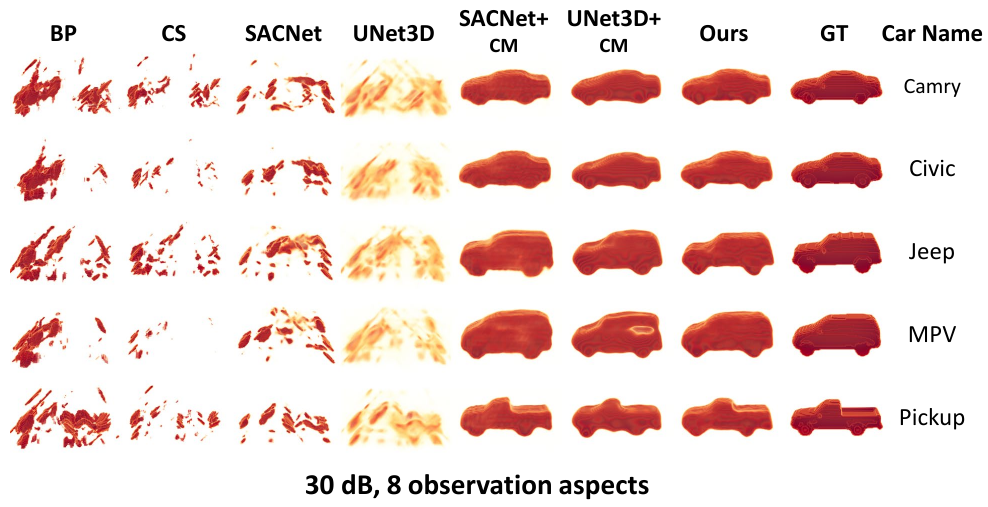}
	\caption{Imaging results of different methods. Each column corresponds to an imaging method, and each row represents a vehicle type. Imaging at an SNR of \textbf{30dB} and \textbf{8} observation aspects.}
	\label{fig:simu2}
\end{figure*}

Figure \ref{fig:simu1} shows the 3D reconstruction results of CMAR-Net under an SNR of 30 dB with 8 observation aspects, together with the corresponding BP imaging results and 3D model visualizations. To better illustrate reconstruction quality, the CMAR-Net output is rendered from three viewpoints, specified by (azimuth, elevation) = (0°, 17°), (30°, 17°), and (135°, 17°). For comparison, the BP result and its corresponding 3D model are visualized at the (0°, 17°) viewpoint, which is also used as the default viewpoint in all subsequent experimental results unless otherwise noted. As shown, the BP method—constrained by the limited number of observation aspects—produces strong sidelobe artifacts and incomplete vehicle structures. In contrast, CMAR-Net markedly improves the reconstruction of the vehicle surface, recovering visually distinguishable geometric features, and the resulting shape closely matches the reference 3D model.

\subsubsection{Comparison with Baseline Methods}

\begin{figure*}[!h]
	\centering
	\includegraphics[width=0.7\linewidth,page=2]{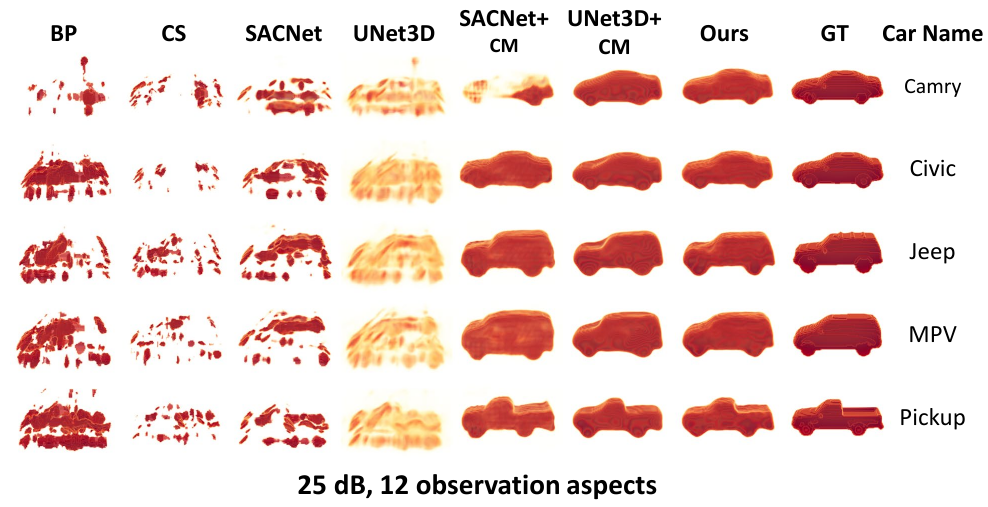}
	\caption{Imaging results of different methods at an SNR of \textbf{25dB} and \textbf{12} observation aspects.}
	\label{fig:simulation-snr25}
\end{figure*}

\begin{figure*}[!h]
	\centering
	\includegraphics[width=0.7\linewidth,page=3]{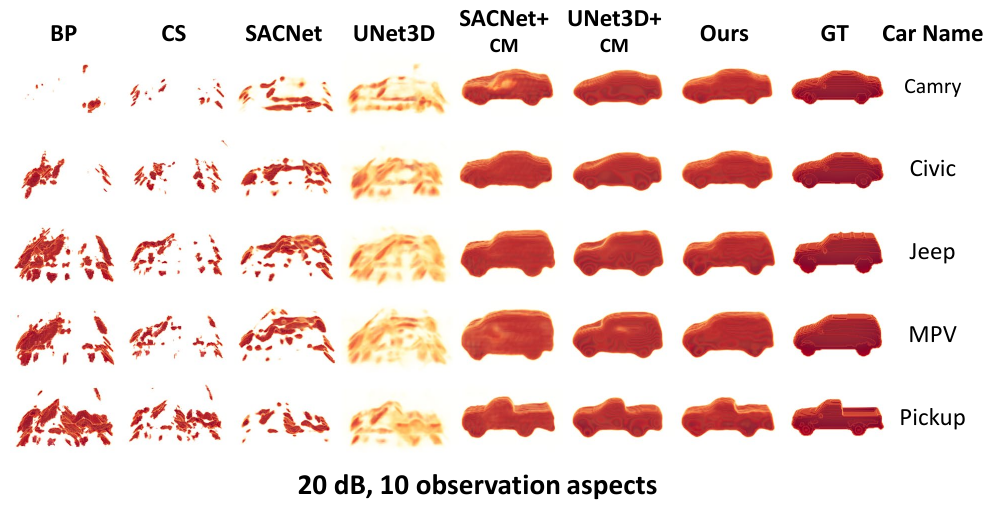}
	\caption{Imaging results of different methods at an SNR of \textbf{20dB} and \textbf{10} observation aspects.}
	\label{fig:simulation-snr20}
\end{figure*}

\begin{figure*}[!h]
	\centering
	\includegraphics[width=0.7\linewidth,page=4]{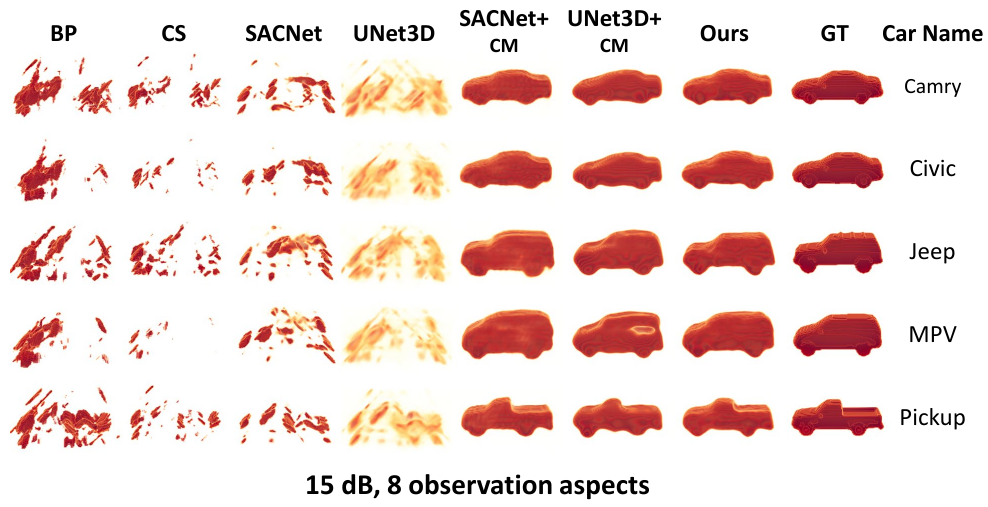}
	\caption{Imaging results of different methods at an SNR of \textbf{15dB} and \textbf{8} observation aspects.}
	\label{fig:simulation-snr15}
\end{figure*}

\begin{figure*}[ht]
	\centering
	\includegraphics[width=0.7\linewidth,page=5]{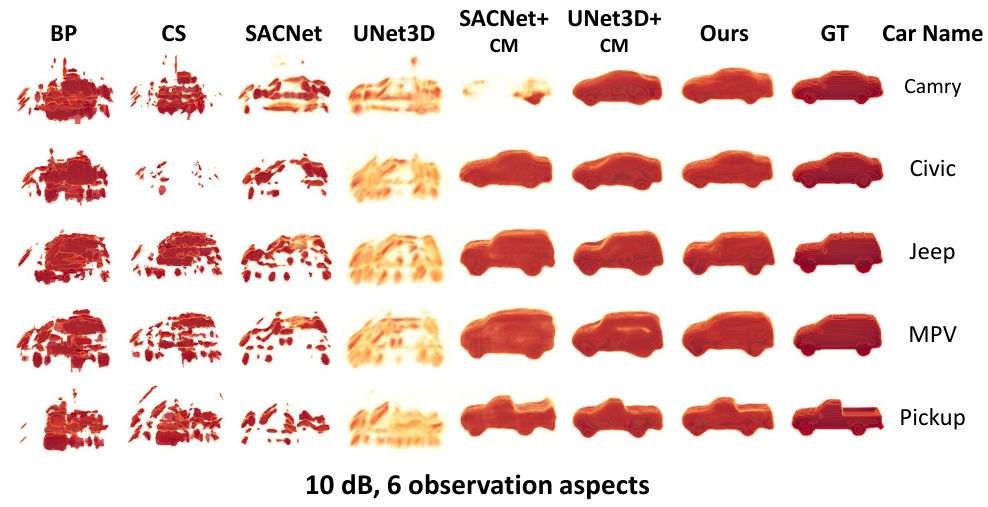}
	\caption{Imaging results of different methods at an SNR of \textbf{10dB} and \textbf{6} observation aspects.}
	\label{fig:simulation-snr10}
\end{figure*}

\begin{figure*}[ht]
	\centering
	\includegraphics[width=0.7\linewidth,page=6]{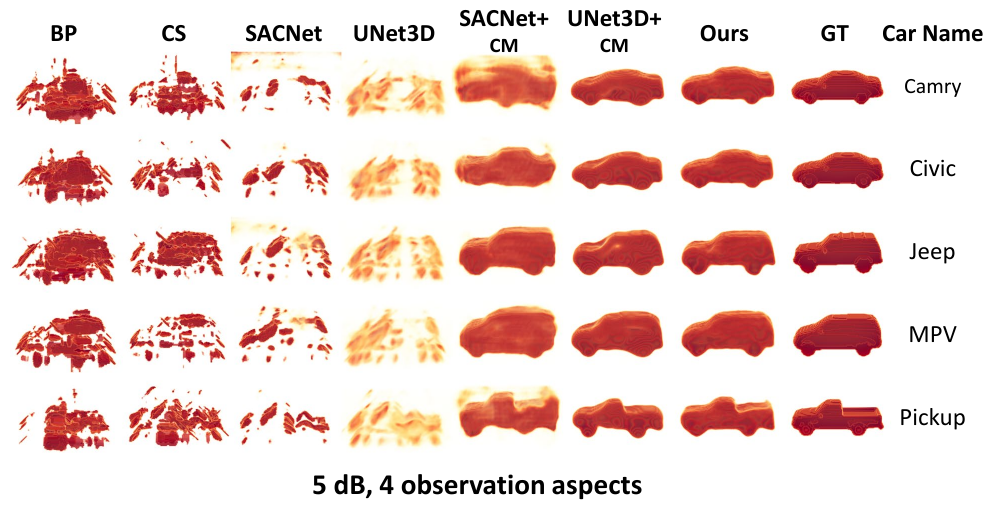}
	\caption{Imaging results of different methods at an SNR of \textbf{5dB} and \textbf{4} observation aspects.}
	\label{fig:simulation-snr5}
\end{figure*}

We compare CMAR-Net with baseline methods. Figure \ref{fig:simu2} presents the imaging results of different methods at an SNR of 30dB and 8 observation aspects. The BP method preserves the strong scattering points of the vehicle body by non-coherently integrating the maximum amplitude values from sub-aspect images. For instance, the bucket structure of Pickup are prominently improved. However, the imaging quality is degraded by high side lobes, as evidenced by the distorted front and windshield of the MPV and Camry. The CS method mitigates the impact of side lobes by optimizing sub-aspect imaging. However, the imaging results reveal significant structural omissions, attributed to the sparsity assumptions and the configuration of the optimization parameters.

DL methods supervised by radar images leverage prior knowledge from full-aspect (360°) imaging, enabling the reconstruction of vehicle images with more scattering details using sparse data. Such as the clearer bucket on the Pickup. However, due to constraints imposed by electromagnetic characteristics, the imaging results lack well-defined planar structures on the vehicle body. 

In contrast, CMAR-Net does not suffer from structural loss and provides accurate geometry and clear vehicle contours, making it the best-performing method among all approaches. For example, in the reconstruction of the MPV, CMAR-Net restores a more complete rear body, while for the Jeep, it presents clearer contours and more precise geometric representation. Additionally, we found that after integrating cross-modal supervision, the performance of SACNet and UNet3D also improved significantly, further validating the applicability of the cross-modal supervision strategy to general networks.

To evaluate the reconstruction performance of CMAR-Net under different conditions, we conduct experiments where the SNR decreased from 25 dB to 5 dB, and the number of observation aspects reduced from 12 to 4. Figures \ref{fig:simulation-snr25} to \ref{fig:simulation-snr5} present the imaging results for all experimental conditions.

As the SNR decreases and the number of observation aspects reduces, the imaging results of the BP and CS methods show more side lobes, leading to reduced vehicle recognizability. SACNet and UNet3D demonstrate strong robustness under deteriorating imaging conditions, providing consistent reconstruction results. After integrating the cross-modal supervision, their performance remains stable. In contrast, CMAR-Net consistently delivers the best imaging performance across all vehicle types. Even under the most challenging conditions (5 dB SNR and 4 observation aspects), CMAR-Net achieves excellent reconstruction, while the performance of other methods significantly deteriorates.

\subsubsection{Quantitative comparison}

\begin{figure}[t]
	\centering
	\includegraphics[width=1\linewidth,page=1]{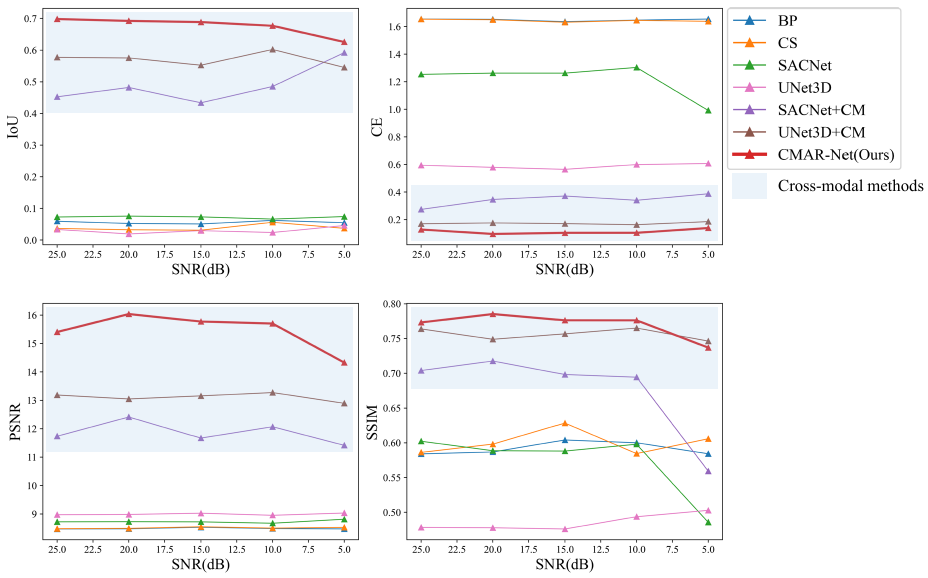}
	\caption{Image quality metrics-SNR curves for different methods on simulated data.}
	\label{fig:cv_metrics_plot}
\end{figure}

We calculate the quality metrics under different data conditions, with the results presented in Tables \ref{tab:simulation_data_IoU} to \ref{tab:simulation_data_SSIM}. Figures \ref{fig:cv_metrics_plot} visualizes these results using curve plots. The data reveal that all existing methods perform worse across all metrics compared to DL-based methods with cross-modal supervision, with particularly notable differences in the IoU and CE metrics. This disparity is largely due to modality differences between the output images and the ground truth data. And among the proposed DL-based methods, CMAR-Net consistently achieves the best performance across all metrics. The only exception occurs under the condition of 4 aspects and an SNR of 5 dB, where CMAR-Net’s SSIM is slightly lower than the method using UNet3D as the backbone. We believe this is due to the UNet3D's tendency to output higher pixel values (closer to 1). However, this behavior in fact leads to a loss of structural information, exhibiting poor generalization of the network, a conclusion that will be confirmed in real-world experiments.

In summary, compared to traditional methods and DL methods supervised with radar images, CMAR-Net achieved improvements of $>$100\% in IoU, 91.23\% in CE, 80.28\% in PSNR and 39.32\% in SSIM. When compared to DL methods with cross-modal supervision, CMAR-Net shows enhancements of 22.54\% in IoU, 52.84\% in CE, 24.32\% in PSNR and 13.09\% in SSIM. These results demonstrate that CMAR-Net can reconstruct more realistic images than other methods.

\begin{table*}[t]
	\centering
	\caption{IoU$\uparrow$/CE$\downarrow$ comparison across all methods on simulated data.}
	\label{tab:simulation_data_IoU}
	\resizebox{\textwidth}{!}{%
		\begin{tabular}{@{}llllllllllll@{}}
			\toprule
			\multicolumn{2}{l}{\multirow{2}{*}{Methods}} &
			\multicolumn{1}{c}{\begin{tabular}[c]{@{}c@{}}Number of\\ observation aspects\end{tabular}} &
			\multicolumn{3}{c}{12} &
			\multicolumn{3}{c}{8} &
			\multicolumn{3}{c}{4} \\ \cmidrule(l){3-12} 
			\multicolumn{2}{l}{}          & SNR & \multicolumn{1}{c}{30dB}  & \multicolumn{1}{c}{15dB}  & \multicolumn{1}{c|}{5dB}   & \multicolumn{1}{c}{30dB}  & \multicolumn{1}{c}{15dB}  & \multicolumn{1}{c|}{5dB}   & \multicolumn{1}{c}{30dB}  & \multicolumn{1}{c}{15dB}  & \multicolumn{1}{c}{5dB}   \\ \midrule
			\multicolumn{2}{l}{BP\cite{dunganCivilianVehicleRadar2010}}        &     & 0.051/1.624 & 0.055/1.685 & \multicolumn{1}{l|}{0.045/1.637} & 0.077/1.625 & 0.067/1.640 & \multicolumn{1}{c|}{0.051/1.636} & 0.064/1.656 & 0.047/1.629 & 0.055/1.654 \\
			\multicolumn{2}{l}{CS\cite{ertinGOTCHAExperienceReport2007}}        &     & 0.032/1.634 & 0.036/1.647 & \multicolumn{1}{l|}{0.031/1.637} & 0.059/1.622 & 0.025/1.631 & \multicolumn{1}{c|}{0.031/1.632} & 0.046/ 1.643 & 0.027/1.640 & 0.037/1.638 \\
			\multicolumn{2}{l}{SACNet\cite{wangMultibaselineSAR3D2023}}    &     & 0.078/1.307 & 0.081/1.294 & \multicolumn{1}{l|}{0.069/1.091} & 0.079/1.357 & 0.075/1.196 & \multicolumn{1}{l|}{0.073/1.263} & 0.075/1.283 & 0.078/1.332 & 0.074/0.992 \\
			\multicolumn{2}{l}{UNet3D\cite{wangSingleTargetSAR2021}}    &     & 0.031/0.491 & 0.030/0.503 & \multicolumn{1}{l|}{0.041/0.593} & 0.027/0.511 & 0.041/0.545 & \multicolumn{1}{l|}{0.030/ 0.565} & 0.038/0.526 & 0.034/0.533 & 0.046/ 0.607 \\ \midrule
			\multicolumn{2}{l}{SACNet+\textbf{CM}} &     & 0.653/0.205 & 0.744/0.171 & \multicolumn{1}{l|}{0.269/0.530} & 0.717/0.215 & 0.755/0.153 & \multicolumn{1}{c|}{0.434/0.371} & 0.558/0.464 & 0.462/0.498 & 0.592/0.387 \\
			\multicolumn{2}{l}{UNet3D+\textbf{CM}} &     & 0.639/0.167 & 0.638/0.174 & \multicolumn{1}{l|}{0.601/0.187} & 0.639/0.154 & 0.604/0.195 & \multicolumn{1}{c|}{0.553/0.171} & 0.593/0.154 & 0.586/0.179 & 0.545/ 0.186 \\
			\multicolumn{2}{l}{CMAR-Net(\textbf{Ours})} &
			&
			\textbf{0.750/0.099} &
			\textbf{0.744/0.108} &
			\textbf{0.713/0.096} &
			\textbf{0.745/0.112} &
			\textbf{0.726/0.109} &
			\textbf{0.689/0.104} &
			\textbf{0.694/0.119} &
			\textbf{0.670/0.111} &
			\textbf{0.626/0.139} \\ \bottomrule
		\end{tabular}%
	}
\end{table*}

\begin{table*}[t]
	\centering
	\caption{SSIM$\uparrow$/PSNR$\uparrow$ comparison across all methods on simulated data.}
	\label{tab:simulation_data_SSIM}
	\resizebox{\textwidth}{!}{%
		\begin{tabular}{@{}llllllllllll@{}}
			\toprule
			\multicolumn{2}{l}{\multirow{2}{*}{Methods}} &
			\multicolumn{1}{c}{\begin{tabular}[c]{@{}c@{}}Number of\\ observation aspects\end{tabular}} &
			\multicolumn{3}{c}{12} &
			\multicolumn{3}{c}{8} &
			\multicolumn{3}{c}{4} \\ \cmidrule(l){3-12} 
			\multicolumn{2}{c}{} &
			SNR &
			\multicolumn{1}{c}{30dB} &
			\multicolumn{1}{c}{15dB} &
			\multicolumn{1}{c|}{5dB} &
			\multicolumn{1}{c}{30dB} &
			\multicolumn{1}{c}{15dB} &
			\multicolumn{1}{c|}{5dB} &
			\multicolumn{1}{c}{30dB} &
			\multicolumn{1}{c}{15dB} &
			\multicolumn{1}{c}{5dB} \\ \midrule
			\multicolumn{2}{l}{BP}        &  & 0.612/8.56 & 0.574/8.39 & \multicolumn{1}{l|}{0.612/8.52} & 0.580/8.56 & 0.584/8.51 & \multicolumn{1}{l|}{0.604/8.53} & 0.574/8.47 & 0.605/8.55 & 0.584/8.48          \\
			\multicolumn{2}{l}{CS}        &  & 0.626/8.54 & 0.603/8.50 & \multicolumn{1}{l|}{0.628/8.53} & 0.591/8.57 & 0.637/8.55 & \multicolumn{1}{l|}{0.628/8.55} & 0.589/8.51 & 0.626/8.53 & 0.606/8.53          \\
			\multicolumn{2}{l}{SACNet}    &  & 0.626/8.80 & 0.613/8.80 & \multicolumn{1}{l|}{0.595/8.80} & 0.615/8.73 & 0.608/8.80 & \multicolumn{1}{l|}{0.588/8.72} & 0.585/8.77 & 0.574/8.74 & 0.486/8.82          \\
			\multicolumn{2}{l}{UNet3D}    &  & 0.447/9.22 & 0.452/9.04 & \multicolumn{1}{l|}{0.484/9.04} & 0.460/9.16 & 0.477/9.15 & \multicolumn{1}{l|}{0.476/9.03} & 0.480/9.18 & 0.457/9.15 & 0.503/9.03          \\ \midrule
			\multicolumn{2}{l}{SACNet+\textbf{CM}} &  & 0.667/13.21 & 0.718/14.54 & \multicolumn{1}{l|}{0.695/10.48} & 0.696/13.74 & 0.724/14.71 & \multicolumn{1}{l|}{0.698/11.67} & 0.486/10.47 & 0.536/10.65 & 0.559/11.42          \\
			\multicolumn{2}{l}{UNet3D+\textbf{CM}} &  & 0.763/13.18 & 0.762/13.14 & \multicolumn{1}{l|}{0.750/12.96} & 0.774/13.41 & 0.744/12.81 & \multicolumn{1}{l|}{0.757/13.16} & 0.766/13.33 & 0.753/12.97 & \textbf{0.746}/12.90 \\
			\multicolumn{2}{l}{CMAR-Net(\textbf{Ours})} &
			&
			\textbf{0.820/16.45} &
			\textbf{0.811/16.21} &
			\textbf{0.796/16.24} &
			\textbf{0.805/16.11} &
			\textbf{0.796/15.82} &
			\textbf{0.776/15.77} &
			\textbf{0.782/15.38} &
			\textbf{0.778/15.45} &
			0.737/\textbf{14.33} \\ \bottomrule
		\end{tabular}%
	}
\end{table*}

\subsubsection{Feature space interpolation}

\begin{figure*}[t]
	\centering
	\includegraphics[width=0.85\linewidth]{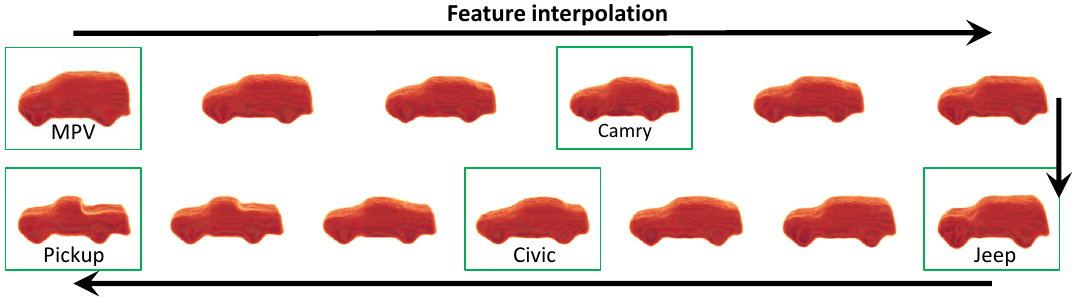}
	\caption{Feature space interpolation results.}
	\label{fig:interpolation}
\end{figure*}

Figure \ref{fig:interpolation} shows the results of the feature space interpolation experiment. In this experiment, we perform linear interpolation between feature vectors extracted by the encoder structure in CMAR-Net, and observe how these interpolated features affect the generated output. The green boxes in the figure represent the output vehicle images of the training type. The images between each pair of green boxes represent the output results of the interpolated feature space. As shown in the figure, linear interpolation between two feature vectors leads to smooth and meaningful nonlinear transitions in the image space\cite{8100128, bojanowski2017optimizing, 9555209}. Notably, the variations in body height and tail features of the interpolated images do not align with any specific vehicle type from the training data. This indicates that the network preserves distinct vehicle features during interpolation, rather than averaging them into an "average" representation\cite{10214202}. We believe this is a superior property of the reconstruction network, as it highlights the model’s ability to generate diverse outputs. This shows the model's flexibility and capability to produce realistic and varied vehicle reconstructions.

\subsection{Results for real-world data}
\subsubsection{Real-world dataset}
\begin{figure*}[t]
	\centering
	\includegraphics[width=.8\linewidth]{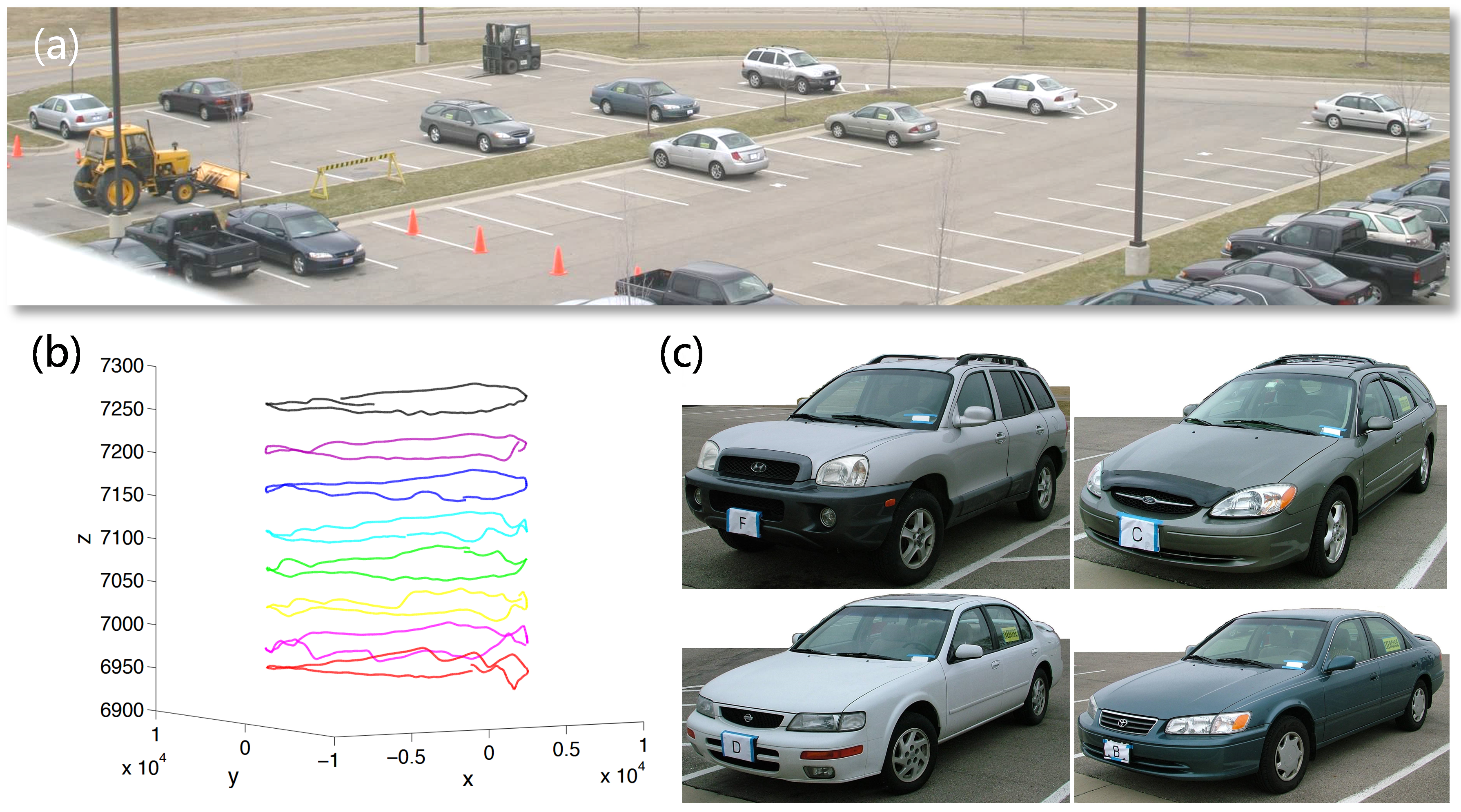}
	\caption{Real-world data description. (a) Data collection scene. (b) Radar flight passes. (c) Photos of test vehicles\cite{casteel2007challenge}.}
	\label{fig:measured-dataset}
\end{figure*}
The real-world data used in this study is sourced from the GOTCHA Circular SAR dataset, collected and released by the United States Air Force Laboratory\cite{casteel2007challenge}. This dataset captures a scene with various civilian vehicles using a radar with the center frequency of 9.6 GHz and the bandwidth of 640 MHz. The data is acquired in circular SAR mode through 8 baselines at different altitudes, with elevation angles of [45.66°, 44.01°, 43.92°, 44.18°, 44.14°, 43.53°, 43.01°, 43.06°]. The combination of circular observation aspect and altitude diversity enabled 3D imaging of the entire scene. We select data from two SUVs and two sedans. Figure \ref{fig:measured-dataset} illustrates the scene, radar flight baselines, and vehicle photos.

\subsubsection{Imaging results}
\begin{figure*}[!t]
	\centering
	\includegraphics[width=.8\linewidth,page=1]{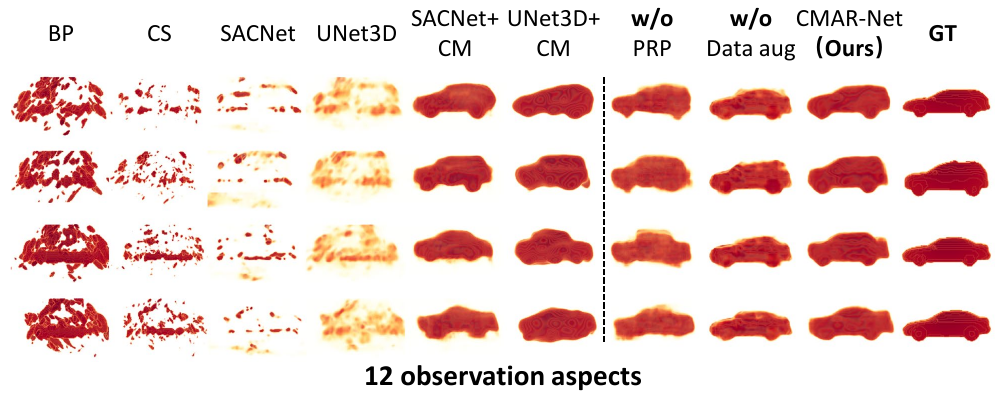}
	\caption{Imaging results of different methods using real-world data. Imaging using \textbf{12} observation aspects}
	\label{fig:measured-results-12}
\end{figure*}

\begin{figure*}[!t]
	\centering
	\includegraphics[width=0.8\linewidth,page=2]{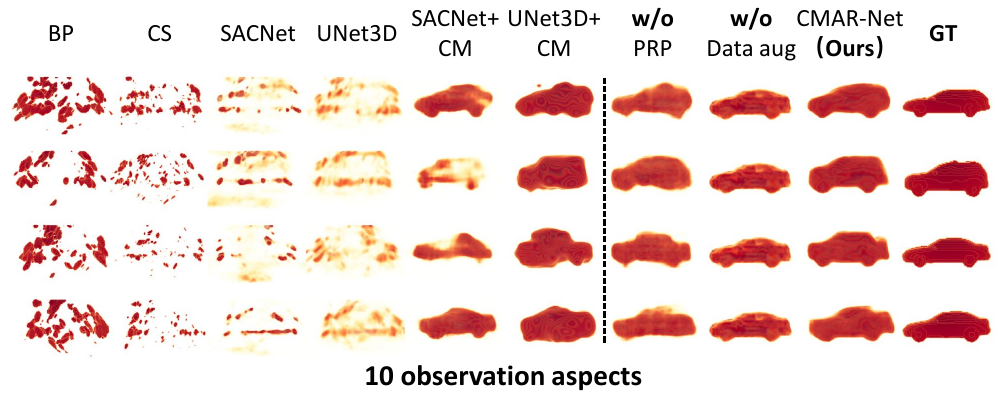}
	\caption{Imaging results of different methods using \textbf{real-world} data. Imaging using \textbf{10} observation aspects}
	\label{fig:measured-results-10}
\end{figure*}

\begin{figure*}[!t]
	\centering
	\includegraphics[width=.8\linewidth,page=3]{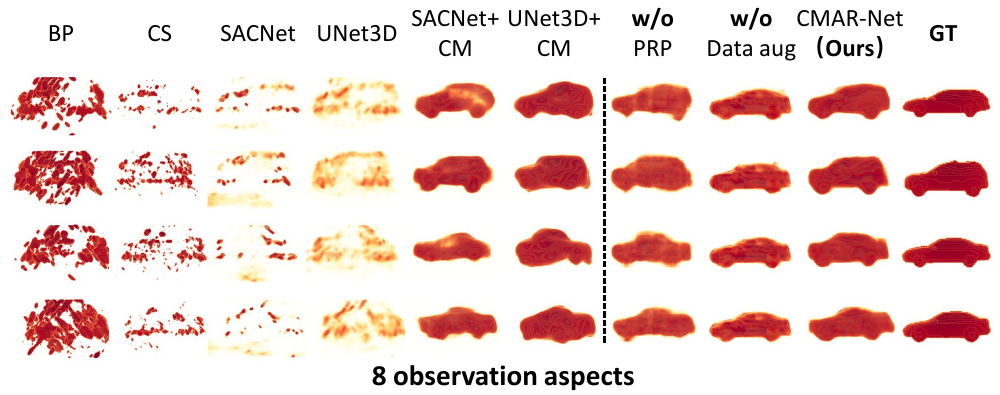}
	\caption{Imaging results of different methods using \textbf{real-world} data. Imaging using \textbf{8} observation aspects}
	\label{fig:measured-results-8}
\end{figure*}

\begin{figure*}[!t]
	\centering
	\includegraphics[width=.8\linewidth,page=4]{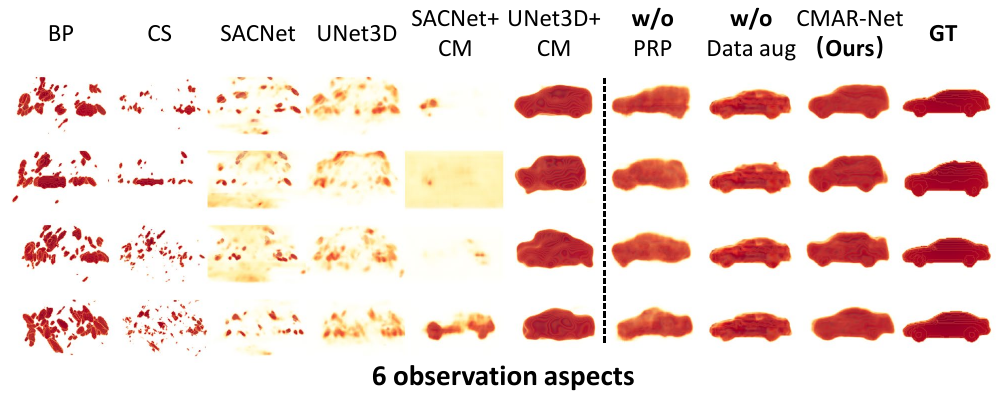}
	\caption{Imaging results of different methods using \textbf{real-world} data. Imaging using \textbf{6} observation aspects}
	\label{fig:measured-results-6}
\end{figure*}

\begin{figure*}[!t]
	\centering
	\includegraphics[width=.8\linewidth,page=5]{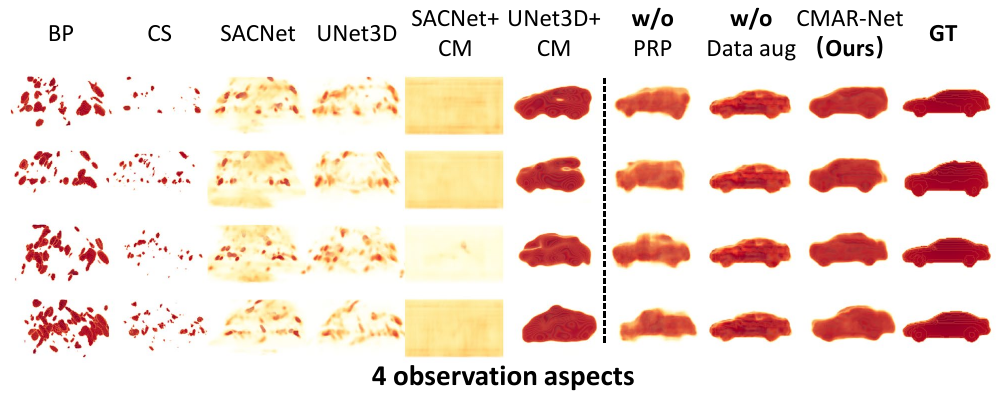}
	\caption{Imaging results of different methods using \textbf{real-world} data. Imaging using \textbf{4} observation aspects}
	\label{fig:measured-results-4}
\end{figure*}
Figure \ref{fig:measured-results-12} to \ref{fig:measured-results-4} present the imaging results on real-world data with varying number of observation aspects. To facilitate observation, the visualization angle is set to (0°, 0°) when presenting the results. The BP method produces results heavily affected by noise and side lobes, which can hardly be recognized. The CS method alleviates these issues, but the reconstructions still appear as chaotic point clouds, leaving vehicle features unrecognizable, especially when using highly sparse data. For example, when the number of observation aspects is 4, traditional methods completely fail, producing totally disordered point clusters without any discernible or meaningful vehicle features.

In contrast, radar image-supervised DL methods can reconstruct vaguely discernible vehicle contours. However, as the number of aspects decreases, these contours deteriorates. For example, when the number of aspects is reduced to 4, SACNet and UNet3D show significant omissions in reconstructing the roof of Jeep. After integrating cross-modal supervision, SACNet and UNet3D exhibit weak generalization when processing real-world data, struggling to accurately reconstruct the vehicle's geometry.  Particularly, when the number of aspects is 4, SACNet fails completely and cannot output any visible features.

\begin{figure}[t]
	\centering
	\includegraphics[width=1\linewidth,page=2]{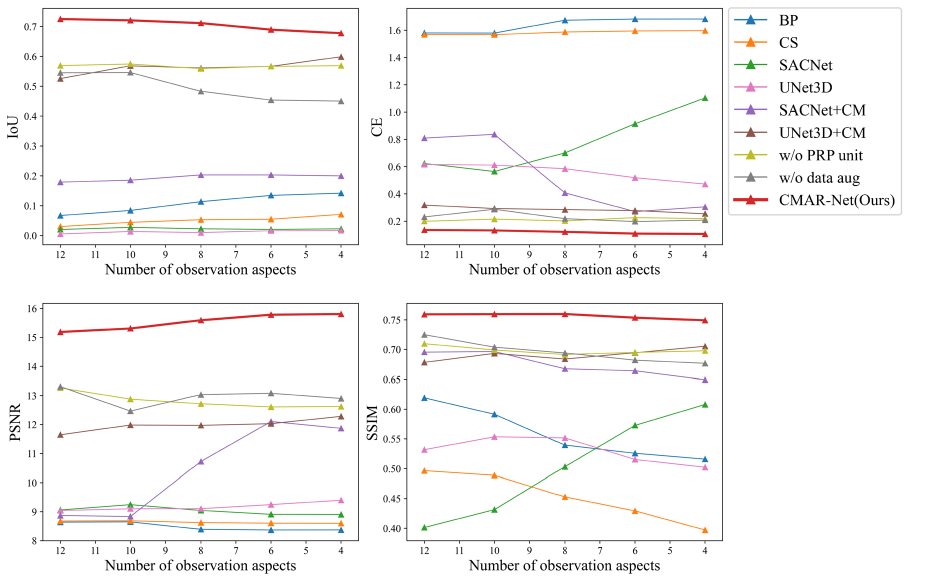}
	\caption{Image quality metrics-number of observation aspects curves for different methods on real-world data.}
	\label{fig:gotcha_metrics_plot}
\end{figure}

Tables \ref{tab:GOTCHA_SSIMPSNR}-\ref{tab:GOTCHA_IoUCE} and Figure \ref{fig:gotcha_metrics_plot} show the evaluation metrics. 
Compared to traditional methods and DL methods supervised with radar images, CMAR-Net achieved improvements of $>$100\% in IoU, 89.36\% in CE, 75.83\% in PSNR and 47.85\% in SSIM. When compared to DL methods with cross-modal supervision, CMAR-Net shows enhancements of 86.42\% in IoU, 67.99\% in CE, 38.81\% in PSNR and 10.75\% in SSIM.

\begin{table}[t]
	\centering
	\caption{SSIM$\uparrow$/PSNR$\uparrow$ comparisons across all methods on real-world data.}
	\label{tab:GOTCHA_SSIMPSNR}
	\resizebox{1\linewidth}{!}{%
		\begin{tabular}{@{}llccccc@{}}
			\toprule
			\multicolumn{2}{l}{\multirow{2}{*}{Methods}} & \multicolumn{5}{c}{Number of observation aspects}                             \\ \cmidrule(l){3-7} 
			\multicolumn{2}{l}{}                         & 12         & 10         & 8          & 6          & 4          \\ \midrule
			\multicolumn{2}{l}{BP}                 & 0.62/8.64  & 0.59/8.65  & 0.54/8.39  & 0.53/8.37  & 0.52/8.37  \\
			\multicolumn{2}{l}{CS}                       & 0.50/8.68  & 0.49/8.69  & 0.45/8.63  & 0.43/8.61  & 0.40/8.60  \\
			\multicolumn{2}{l}{SACNet}                   & 0.40/9.07  & 0.43/9.25  & 0.50/9.05  & 0.57/8.91  & 0.61/8.91  \\
			\multicolumn{2}{l}{UNet3D}                   & 0.53/9.04  & 0.55/9.11  & 0.55/9.11  & 0.52/9.25  & 0.50/9.40  \\ \midrule
			\multicolumn{2}{l}{SACNet+\textbf{CM}}                & 0.70/8.87  & 0.70/8.84  & 0.67/10.73 & 0.66/12.11 & 0.65/11.87 \\
			\multicolumn{2}{l}{UNet3D+\textbf{CM}}                & 0.68/11.65 & 0.69/11.98 & 0.68/11.97 & 0.69/12.03 & 0.71/12.29 \\ \midrule
			\multicolumn{2}{l}{\textbf{w/o} PRP unit}             & 0.71/13.27 & 0.70/12.88 & 0.69/12.72 & 0.70/12.61 & 0.70/12.63 \\
			\multicolumn{2}{l}{\textbf{w/o} data aug}             & 0.72/13.31 & 0.70/12.47 & 0.69/13.04 & 0.68/13.08 & 0.68/12.90 \\ \midrule
			\multicolumn{2}{l}{CMAR-Net(\textbf{Ours})} & \textbf{0.76/15.19} & \textbf{0.76/15.31} & \textbf{0.76/15.59} & \textbf{0.75/15.78} & \textbf{0.75/15.81} \\ \bottomrule
		\end{tabular}%
	}
\end{table}

\begin{table}[t]
	\centering
	\caption{IoU$\uparrow$/CE$\downarrow$ comparisons across all methods on real-world data.}
	\label{tab:GOTCHA_IoUCE}
	\resizebox{1\linewidth}{!}{%
		\begin{tabular}{@{}llccccc@{}}
			\toprule
			\multicolumn{2}{l}{\multirow{2}{*}{Methods}} & \multicolumn{5}{c}{Number of observation aspects}                        \\ \cmidrule(l){3-7} 
			\multicolumn{2}{l}{}                         & 12        & 10        & 8         & 6         & 4         \\ \midrule
			\multicolumn{2}{l}{BP}                 & 0.07/1.58 & 0.08/1.58 & 0.11/1.67 & 0.13/1.68 & 0.14/1.68 \\
			\multicolumn{2}{l}{CS}                       & 0.03/1.57 & 0.04/1.57 & 0.05/1.59 & 0.05/1.60 & 0.07/1.60 \\
			\multicolumn{2}{l}{SACNet}                   & 0.02/0.62 & 0.03/0.56 & 0.02/0.70 & 0.02/0.91 & 0.02/1.10 \\
			\multicolumn{2}{l}{UNet3D}                   & 0.01/0.62 & 0.01/0.61 & 0.01/0.58 & 0.02/0.52 & 0.02/0.47 \\ \midrule
			\multicolumn{2}{l}{SACNet+\textbf{CM}}                & 0.18/0.81 & 0.19/0.84 & 0.20/0.41 & 0.20/0.27 & 0.20/0.30 \\
			\multicolumn{2}{l}{UNet3D+\textbf{CM}}                & 0.53/0.32 & 0.57/0.29 & 0.56/0.28 & 0.57/0.28 & 0.60/0.25 \\ \midrule
			\multicolumn{2}{l}{\textbf{w/o} PRP unit}             & 0.57/0.20 & 0.57/0.22 & 0.56/0.20 & 0.57/0.23 & 0.57/0.22 \\
			\multicolumn{2}{l}{\textbf{w/o} data aug}             & 0.55/0.23 & 0.55/0.29 & 0.48/0.22 & 0.45/0.20 & 0.45/0.21 \\ \midrule
			\multicolumn{2}{l}{CMAR-Net(\textbf{Ours})} & \textbf{0.72/0.14} & \textbf{0.72/0.13} & \textbf{0.71/0.12} & \textbf{0.69/0.11} & \textbf{0.68/0.11} \\ \bottomrule
		\end{tabular}%
	}
\end{table}
Moreover, compared to simulated data, other DL methods experience a significant decline in performance, with an average drop of 12.17\% across all metrics. In contrast, CMAR-Net maintains a high score, with only a 4.66\% drop, significantly outperforming other methods and demonstrating its superior generalization ability. This is attributed to the PRP unit and data augmentation strategy, as further validated by the ablation experiments.

\subsubsection{Ablation study}
The 7th and 8th sets of data in the Figure \ref{fig:measured-results-12}-\ref{fig:measured-results-4} and table \ref{tab:GOTCHA_SSIMPSNR}-\ref{tab:GOTCHA_IoUCE} show the imaging results and evaluation metrics after ablating the PRP unit and data augmentation strategy. Without the PRP unit, the vehicle images become blurry and suffer greater pixel loss as the number of aspects decreases. And in the absence of data augmentation, the network consistently reconstructs all vehicle type as sedan type, revealing severe overfitting. The PRP unit design improves SSIM, PSNR, IoU, and CE by 8.25\%, 21.25\%, 24.15\%, and 42.82\%, respectively. Data augmentation contributes improvements of 8.65\%, 19.92\%, 42.94\%, and 46.75\%. These results clearly demonstrate the effectiveness of both the PRP unit and data augmentation strategy.

\section{Conclusion}
\label{conclusion}
In this paper, we propose a DL-based sparse 3D SAR reconstruction method for vehicle target. We introduce the concept of cross-modal learning and design a DNN tailored for 3D SAR reconstruction—CMAR-Net. By leveraging differentiable rendering, CMAR-Net can be effectively trained, addressing the key challenge of supervising 3D reconstruction network with 2D optical images. Extensive experiments on both simulated and real-world datasets show that CMAR-Net significantly outperforms all existing baseline methods, including traditional imaging algorithms and DL-based approaches, in terms of image quality and generalization. Moreover, the existing DL-based SAR reconstruction methods, adopting our proposed cross-modal learning paradigm, experience a significant performance improvement, which highlights the superiority and versatility. Last but not least, our method eliminates the need for time-consuming preprocessing of full-aspect data. Instead, it only needs 2D images obtained through computer rendering or camera capture. This significantly reduces the difficulty of dataset construction, greatly enhancing the practicality of DL-based 3D SAR reconstruction technology.


Our approach provides new insights into 3D SAR research, particularly advancing target imaging. By addressing key challenges in imaging quality, it can support improved applications of 3D SAR in aerial remote sensing, target interpretation, and urban surveillance. Future research could extend this framework to other complex targets, such as buildings or urban environments, and explore integration with additional modalities, such as infrared or hyperspectral data, to further enhance reconstruction performance.

%

\bibliography{BIB}
\bibliographystyle{IEEEtran}

\vfill

\end{document}